%% file: main.tex
\documentclass{article}
\usepackage{spconf,amsmath,graphicx}
\usepackage{color}
\usepackage{booktabs}
\usepackage{multirow}
\usepackage{subfigure}

\title{Enhance transferability of adversarial examples with model architecture}
\name{Mingyuan~Fan$^1$,~Wenzhong~Guo$^1$,~Shengxing~Yu$^3$,~Zuobin~Ying$^2$,~Ximeng~Liu$^1$\thanks{Thanks to the National Natural Science Foundation of China (No.62072109, No.U1804263) for funding.}}
\address{$^1$College of Computer and Data Science, Fuzhou University, China\\
$^2$Faculty of Data Science, City University of Macau, Macau\\
$^3$School of Electronics Engineering and Computer Science, Peking University, China
}

\begin{document}
%
\maketitle
\begin{abstract}
Transferability of adversarial examples is of critical importance to launch black-box adversarial attacks, where attackers are only allowed to access the output of the target model.
However, under such a challenging but practical setting, the crafted adversarial examples are always prone to overfitting to the proxy model employed, presenting poor transferability.
In this paper, we suggest alleviating the overfitting issue from a novel perspective, i.e., designing a fitted model architecture.
Specifically, delving the bottom of the cause of poor transferability, we arguably decompose and reconstruct the existing model architecture into an effective model architecture, namely multi-track model architecture (MMA).
The adversarial examples crafted on the MMA can maximumly relieve the effect of model-specified features to it and toward the vulnerable directions adopted by diverse architectures.
Extensive experimental evaluation demonstrates that the transferability of adversarial examples based on the MMA significantly surpass other state-of-the-art model architectures by up to 40\% with comparable overhead.
\end{abstract}
\section{Introduction}
\input{tex/intro}

\section{Approach}
\input{tex/approach}

\vspace{-0.2cm}
\section{Experiment}
\input{tex/experiment}
\vspace{-0.2cm}

\vspace{-0.1cm}
\section{Conclusion}
\input{tex/conclusion}

\bibliographystyle{IEEEbib}
\bibliography{reference}

\end{document}

%% file: tex/intro.tex
In the past decade, deep neural networks (DNNs) have obtained impressive performance on a broad spectrum of real-world applications \cite{cv,healthcare,selfdriving}.
However, superior performance is achieved at the expense of ill robustness to adversarial examples (AEs) \cite{FGSM,pgd,CW}.
Specifically, by only imposing human-imperceptible crafted adversarial noises on legitimate inputs, the resultant AEs exceptionally incur misclassification with high confidence from advanced DNNs.
It sparks increasing concerns about the reliability of DNNs and significantly hinders the applications of DNNs in practice, especially for security-critical tasks, e.g., self-driving \cite{adv_for_selfdriving1,adv_for_selfdriving2}.
\par
AEs can be utilized to spot the blind aspect of DNNs, and hence adversarial attacks become the fundamental task in the AI secure domain.
Generally, there are two kinds of approaches adopted to generate AEs, namely white-box attacks \cite{CW,auto_pgd,composite_attack} and black-box attacks \cite{DaST,bayes_attack,hybrid_batch_attack}.
For white-box attacks, the attackers have full access to the target model, e.g., model architecture; whereas, in the case of black-box attacks, only predictions of the model to inputs are available to attackers.
Due to the practicality of the black-box scenario, black-box attacks are more challenging and valuable.
\par
There are two feasible solutions to accomplish black-box attacks, i.e., query-based attacks \cite{hybrid_batch_attack,bayes_attack} and transfer-based attacks \cite{feature_space_trans,PID}.
Query-based attack builds in estimating the gradient of the target model to input and has to require huge queries to the target model, e.g., hundreds per image in CIFAR-10.
The anomaly query probably raises the doubt of the model-owner, leading to the prohibition of the attacker's behavior.
On the contrary, transfer-based attacks desire to train a proxy model for simulating the target model and then harness the proxy model to produce AEs, without the interaction to the target model.
Therefore, transfer-based attacks have more practical applicability than query-based attacks.
\par
The effectiveness of transfer-based attacks totally depends on the cross-model transferability of generated AEs.
Furthermore, the transferability stems from the shared low-level features over different models \cite{adv_is_feature}.
However, the AEs generated by vanilla transfer-based attacks (i.e., adopting white-box attacks on the proxy model) are indeed prone to overfit the exclusive high-level features of the proxy model, ignoring the shared low-level features.
For example, the crafted AEs achieve the 100\% success rates in the proxy model, but the scant success rates (less than 30\% on CIFAR-10) in the target model.
\par
This paper focuses on promoting the transferability of AEs.
Specifically, existing related technologies are done following these aspects \cite{attention,shaply,PID,ghost_net}: robust optimizer, transforming input, ensemble model, or better loss function.
It is ignored whether increasing the transferability can be achieved by designing a proper model architecture, and it inspires us to explore such model architecture.
To do so, we deeply analyze the shortcoming of existing architecture and build on it, design a bespoke model architecture, named multi-track network architecture (MNA), where the AEs based on MNA will towards the shared vulnerable directions over diverse models, greatly ameliorating the overfitting ill.
Besides, the MNA is a generic, not sole, technology of increasing transferability of AEs, and it can be integrated with other such technologies to enhance transferability further.
Extensive experiments show that MNA can improve on transferability of AEs by up to about 40\% without any additional burdens.

%% file: tex/approach.tex
\vspace{-0.2cm}
\subsection{Theoretical Analysis}
\vspace{-0.2cm}
Given an well-trained black-box model $g(\cdot)$, the target is crafting adversarial example $x_{adv}$ for $x$ (with ground-truth label $y$), as well the expection that $x_{adv}$ is closer $x$ as possible.
Due to the limited knowledge about the black-box model $g(\cdot)$, a proxy model $f(\cdot)$ is trained to simulating the black-box model.
Subsequently, the attackers can craft $x_{adv}$ on the proxy model with white-box attacks as follows:
\begin{equation}
\begin{split}
x_{adv} = {Proj}_{x,\epsilon}\{x + \nabla_{x} L(f(x), y)\},
\end{split}
\end{equation}
where ${Proj}_{x,\epsilon}$ is project function that is used to project the input into the $\epsilon$-ball around $x$, $L(\cdot,\cdot)$ is a common loss function, and $\epsilon$ is given by attackers to control the similar between $x_{adv}$ and $x$.
It is noted that the above process generally repeats multiple times.
But, to simplify the deductions and symbols, we employ a single-time process to analyze, and the conclusion can be easily generalized to the iteration form.
\par
Apparently, attackers are desired to make $f(x)=g(x)$, or $\nabla_{x} L(f(x), y) = \nabla_{x} L(g(x), y)$, as possible, thereby achieving the optimal (white-box) attack performance.
Unfortunately, there is a significant divergence between the proxy model and the black-box model due to their own unique architecture, resulting in $f(x) \neq g(x)$.
Next, we explore the source of the divergence and utilize it to guide designing the model architecture with minimal model-cross discrepancy.
Without loss of generality, it is supposed that $f$ and $g$ have the same number of layers to simplify the deduction.
Let $f=f_l \circ f_{l-1} \cdots \circ f_1$  and $g=g_l \circ g_{l-1} \cdots \circ g_1$ where $f_{i}$ and $g_{i}$ can be regarded as i-th layer of $f$ and $g$ and $\circ$ denotes the composite operation, and then there is:
\begin{small}
\begin{equation}
\begin{split}
\nabla_{x}L(f(x),y) = \frac{\partial L}{\partial O^{proxy}_l} \cdot \frac{\partial O^{proxy}_l}{\partial O^{proxy}_{l-1}} \cdots \frac{\partial O^{proxy}_1}{\partial x},
\end{split}
\label{proxy_equ}
\end{equation}
\end{small}
\vspace{-0.2cm}
\begin{small}
\begin{equation}
\begin{split}
\nabla_{x}L(g(x),y) = \frac{\partial L}{\partial O^{black}_l} \cdot \frac{\partial O^{black}_l}{\partial O^{black}_{l-1}} \cdots \frac{\partial O^{black}_1}{\partial x},
\end{split}
\label{black_equ}
\end{equation}
\end{small}
where $O_i^{proxy}$ and $O_i^{black}$ is the output of i-th layer of $f$ and $g$ to $x$.
Note that $\nabla_{x}L(f(x),y)=\nabla_{x}L(g(x),y)$ is established if the each term in the Equation \ref{proxy_equ} is exactly equal to the corresponding term in the Equation \ref{black_equ}.
Therefore, the divergence can be considered as the accumulation of the discrepancy between corresponding terms in Equation \ref{proxy_equ} and Equation \ref{black_equ}.
Besides, it is well-known that DNNs enjoy identical or similar low-level feature extractors, which suggests $f_i \approx g_i$ when $i$ is small than a certain number.
Moreover, the high-level feature extractors that different models learned are always model-specified.
Hence, the poor transferability is heavily induced by the divergence between the late layers of the proxy model and the black-box model.
Next, we will elaborate on how to relieve the problem.

\begin{figure}[t]
    \centering
    \includegraphics[width=0.5\textwidth]{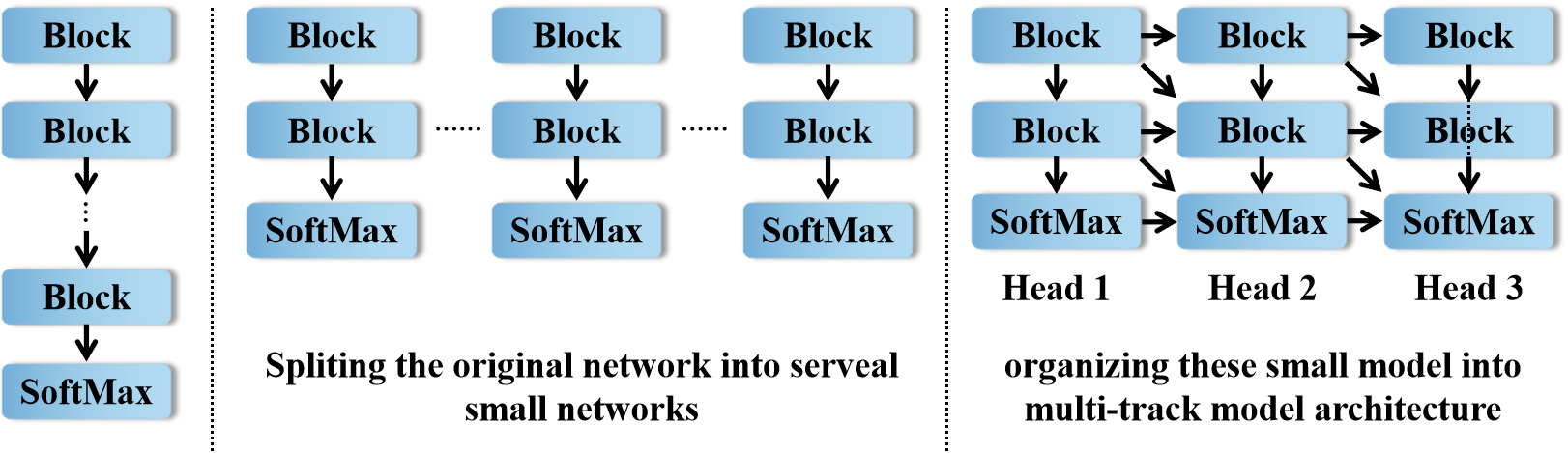}
    \caption{sketch map of MMA construction.}
    \label{model_process}
\end{figure}
\vspace{-0.2cm}
\subsection{Multi-Track model architecture}
\vspace{-0.2cm}
For the convenience of discussion, let
\begin{equation}
\begin{split}
\nabla_{x}L(f(x),y) = f_{high} \cdot f_{low},
\\
\nabla_{x}L(g(x),y) = g_{high} \cdot g_{low},
\end{split}
\end{equation}
where $f_{high},f_{low}$ and $g_{high},g_{low}$ correspond to the high-level and low-level of $f$ and $g$ derivative vectors.
The desired goal is making $f_{high}=g_{high}$, but it is difficult to navigate under the condition without any information about the black-box model.
Therefore, the only leaving solution is increasing the domination of $f_{low}$ in $\nabla_{x}L(f(x),y)$ so that the nagative impact of $f_{high}$ can be narrowed.
However, before presenting our solution, we have to highlight that, excessively increasing the domination of $f_{low}$ does not yield growing benefits to the transferability of AEs.
Simply speaking, the model almost only relies on low-level features when $f_{low}$ dominates $\nabla_{x}L(f(x),y)$.
Furthermore, it implies the poor performance of the model\footnote{Overwhelming successes of deep learning are established by its outstanding capacity for learning high-level semantic features.}, and thus the model does not learn the same low-level feature extractors as the black-box model, i.e., $f_{low}=g_{low}$, which breaking the supposed precondition.
In short, the optimal solution is maintaining the performance of the model while removing as high-level feature extractors as possible.
\par
With the objective of shrinking the negative impact from late layers, the most intuitive solution is evenly splitting the original network into a handful of small networks (as shown in Figure \ref{model_process}).
These small networks have fewer high-level layers than the original network, and thereby low-level feature extractors enjoy a greater proportion in the entire network. 
Unfortunately, the solution suffers from two serious drawbacks.
On the one hand, induced by small model capacity, small networks have to perform poorly, which suggests that the crafting AEs are with ill transferability according to before conclusion.
On the other hand, due to without any prior knowledge, it is difficult to directly determine the exact size of the small models with the best attack performance.
The potential manner is manually-tuning (enumeration, a brute-force method) which still is pretty cumbersome and maybe unbearable to its cost.
The drawbacks produce the quandary of how to achieve the optimal balance between learning low-level feature extractors well and decreasing the disturbance of high-level feature extractors.
\par
To solve the dilemma, we integrate these small models into multi-track model architecture, which can adaptively adjust the model size with comparable overhead to the original network.
In MMA, small models is no longer a separate network from each other; on the contrary, as shown in Figure \ref{model_process}, the network placed in the back is allowed to enjoy the feature maps of the networks placed in the front.
In this way, each small model (in MAA) essentially represents a trade-off point between learning low-level shared features and introducing the disturbance of high-level exclusive features.
Then, the small model with the best attack performance or the entire model can be adopted as the proxy model, rather than manually and carefully adjusting the size of the proxy network and retraining it, which is fairly cumbersome and inelegant.
Finally, we highlight the followed merits of MMA compared to the original network: (1) MMA is a simple yet effective model architecture designing where AEs are more transferable by adaptively adjusting the size of the model, with comparable overhead to the original network; (2) MMA is a generic technology that can be integrated into any existing technology to further enhance the transferability of AEs.


%% file: tex/experiment.tex
\subsection{Experiment Setup}
We examine the performance of MMA with different sizes in CIFAR-10, compared with three widely-used baseline model architectures, namely MobileNet, GoogleNet, and ResNet18.
All proxy models trained for 30 epoch using Momentum optimizer with 0.01 learning rate, 0.9 cumulative factor, and $1 \times 10^{-4}~L_2$ weight decay in the training set of CIFAR-10.
For attack setting, we set $\epsilon$ to 0.1 and employ FGSM (one-step attack) with step size 0.1 and BIM (iteration version of FGSM) with 10 iterations and 0.01 step size ($= \frac{\epsilon}{10}$).
Moreover, MobileNet, GoogleNet, and ResNet18 were also adopted as black-box models for attacking.
To quantify the transferability of resultant AEs, we adopt attack success rate (ASR), the misclassified rate of AEs on the target model, as the metric.
Besides, as shown in Figure \ref{model_process}, the head $i$ denotes the $i$-th output of the MMA and denotes the ensemble output of all outputs of the MAA.
\begin{table}[!h]
\centering
\caption{Transfer results of single-Step attacks.}
\resizebox{0.5\textwidth}{!}{%
\begin{tabular}{@{}c|ccc|ccc@{}}
\toprule
\multirow{2}{*}{Proxy Model} & \multicolumn{3}{c|}{Fixed hyperparameters} & \multicolumn{3}{c}{Tuned (Best) results} \\ \cmidrule(l){2-7} 
 & GoogleNet & MobileNet & ResNet18 & GoogleNet & MobileNet & ResNet18 \\ \midrule
MMA (ours) & \textbf{58.07} & \textbf{47.36} & \textbf{50.80} & \textbf{59.33} & \textbf{54.22} & \textbf{53.10} \\
GoogleNet & - & 37.11 & 45.81 & - & 47.06 & 50.78 \\
MobileNet & 40.11 & - & 40.46 & 42.21 & - & 42.77 \\
ResNet18 & 47.60 & 40.49 & - & 47.60 & 42.00 & - \\ \bottomrule
\end{tabular}
}
\label{singlestep_comp}
\end{table}
\begin{table}[!h]
\centering
\caption{Transfer results of multi-step attacks.}
\resizebox{0.5\textwidth}{!}{%
\begin{tabular}{@{}c|ccc|ccc@{}}
\toprule
\multirow{2}{*}{Proxy Model} & \multicolumn{3}{c|}{Fixed hyperparameters} & \multicolumn{3}{c}{Tuned (Best) results} \\ \cmidrule(l){2-7} 
 & GoogleNet & MobileNet & ResNet18 & GoogleNet & MobileNet & ResNet18 \\ \midrule
MMA (ours) & \textbf{78.50} & \textbf{56.12} & \textbf{65.72} & \textbf{79.79} & \textbf{69.64} & \textbf{70.49} \\
GoogleNet & - & 39.16 & 58.38 & - & 54.87 & 65.10 \\
MobileNet & 41.71 & - & 41.70 & 45.76 & - & 46.50 \\
ResNet18 & 70.58 & 52.80 & - & 70.58 & 55.25 & - \\ \bottomrule
\end{tabular}
}
\label{multistep_comp}
\end{table}
\vspace{-0.2cm}
\subsection{Performance Comparison}
\vspace{-0.2cm}
\begin{figure}[!h]
    \centering
    \subfigure[GoogleNet]{\includegraphics[width=0.32\linewidth]{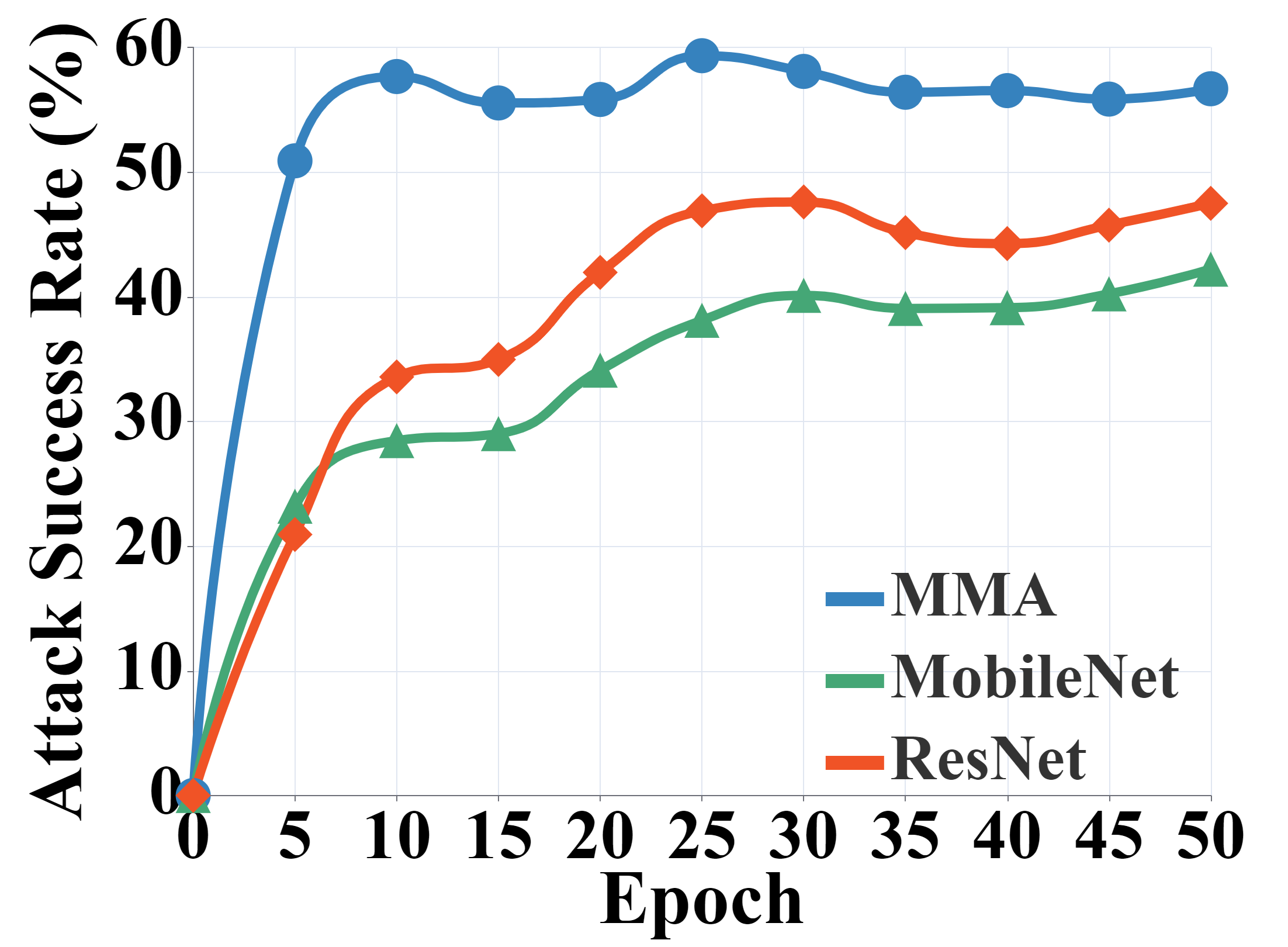}}
    \subfigure[MobileNet]{\includegraphics[width=0.32\linewidth]{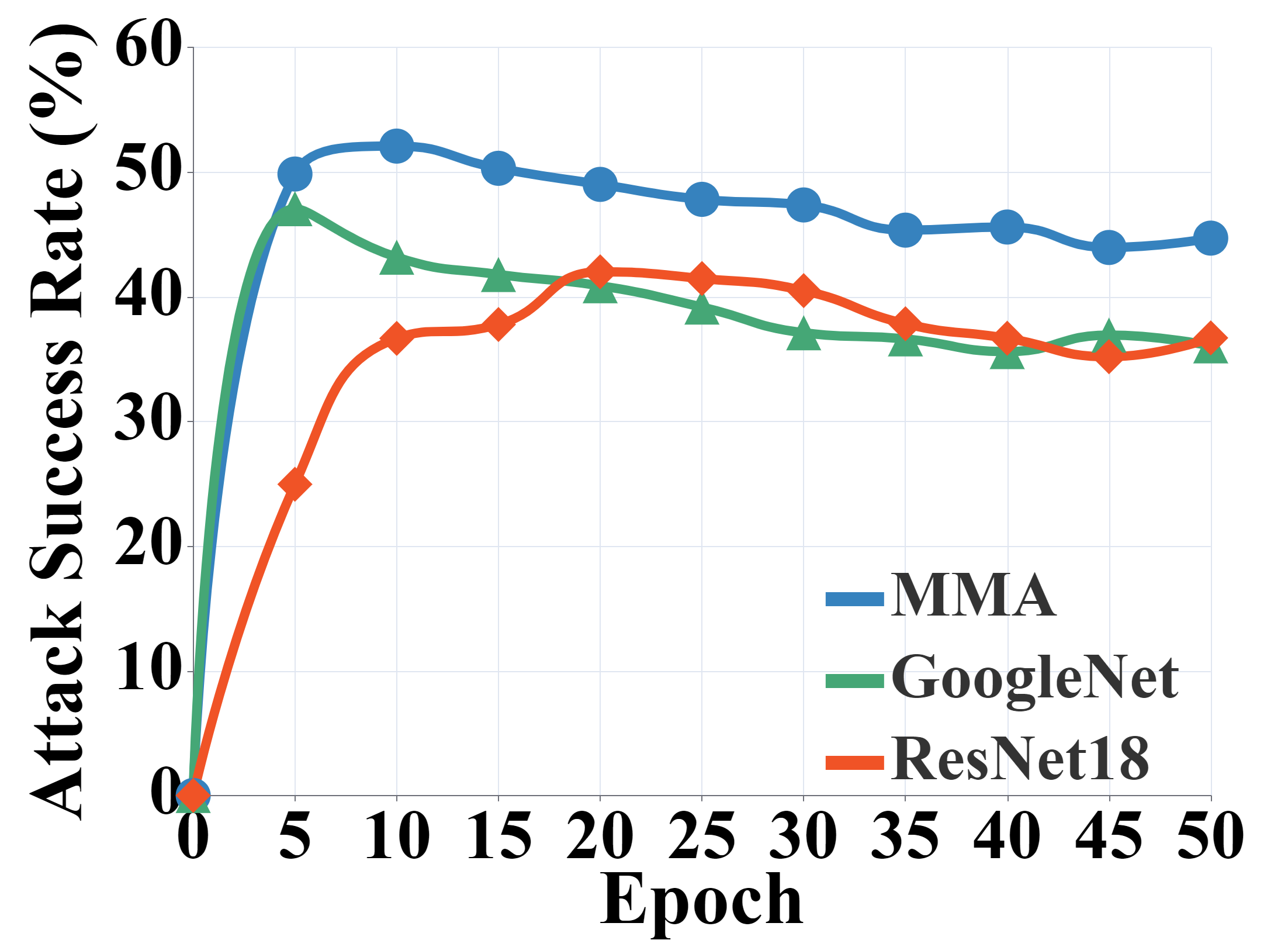}}
    \subfigure[ResNet]{\includegraphics[width=0.32\linewidth]{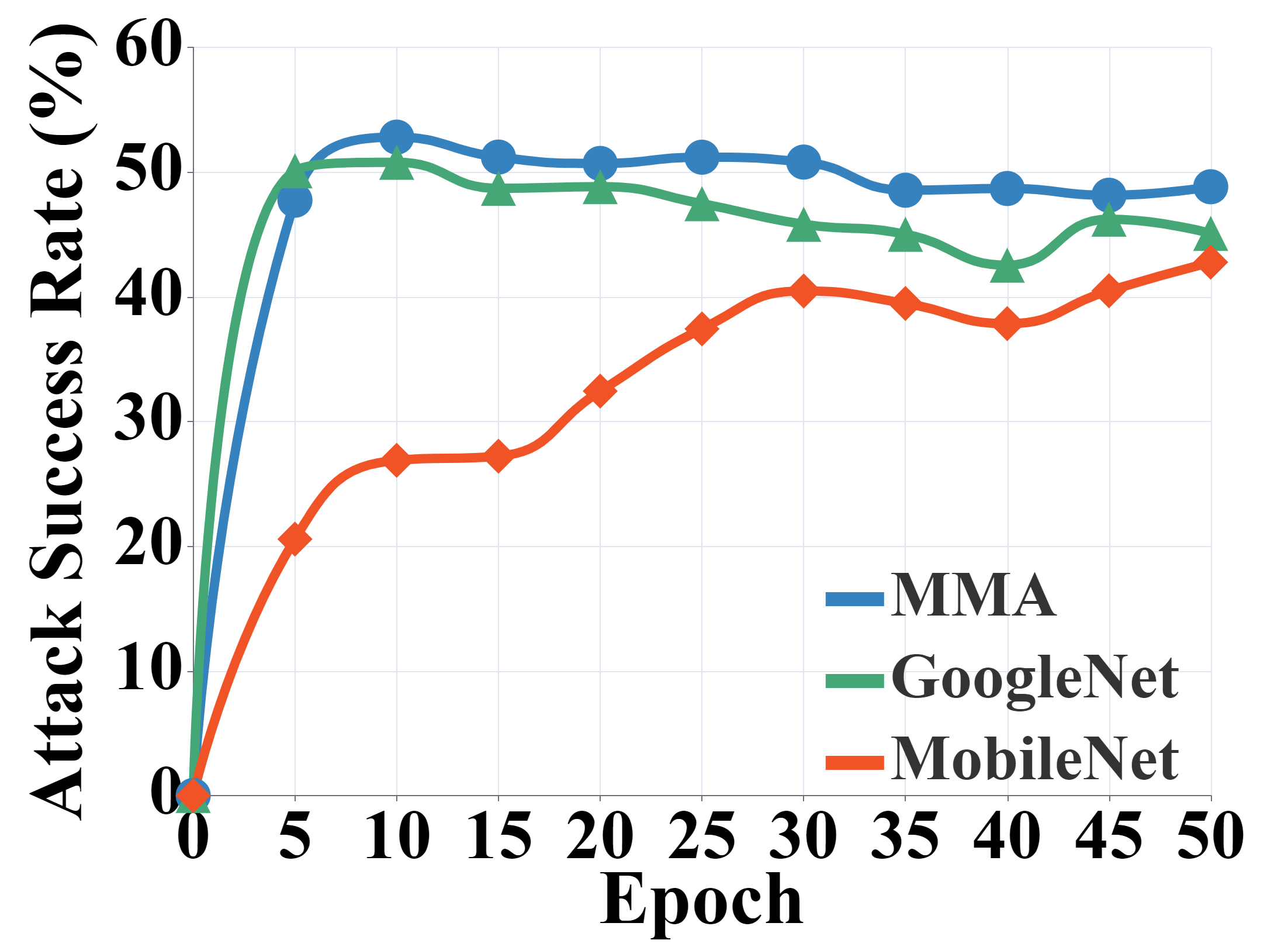}}
    \caption{Single-step attack performances over training epochs. The size of MAA is fixed to $3 \times 4$.}
    \label{epoch_for_single}
\end{figure}

\begin{figure}[!h]
    \centering
    \subfigure[GoogleNet]{\includegraphics[width=0.32\linewidth]{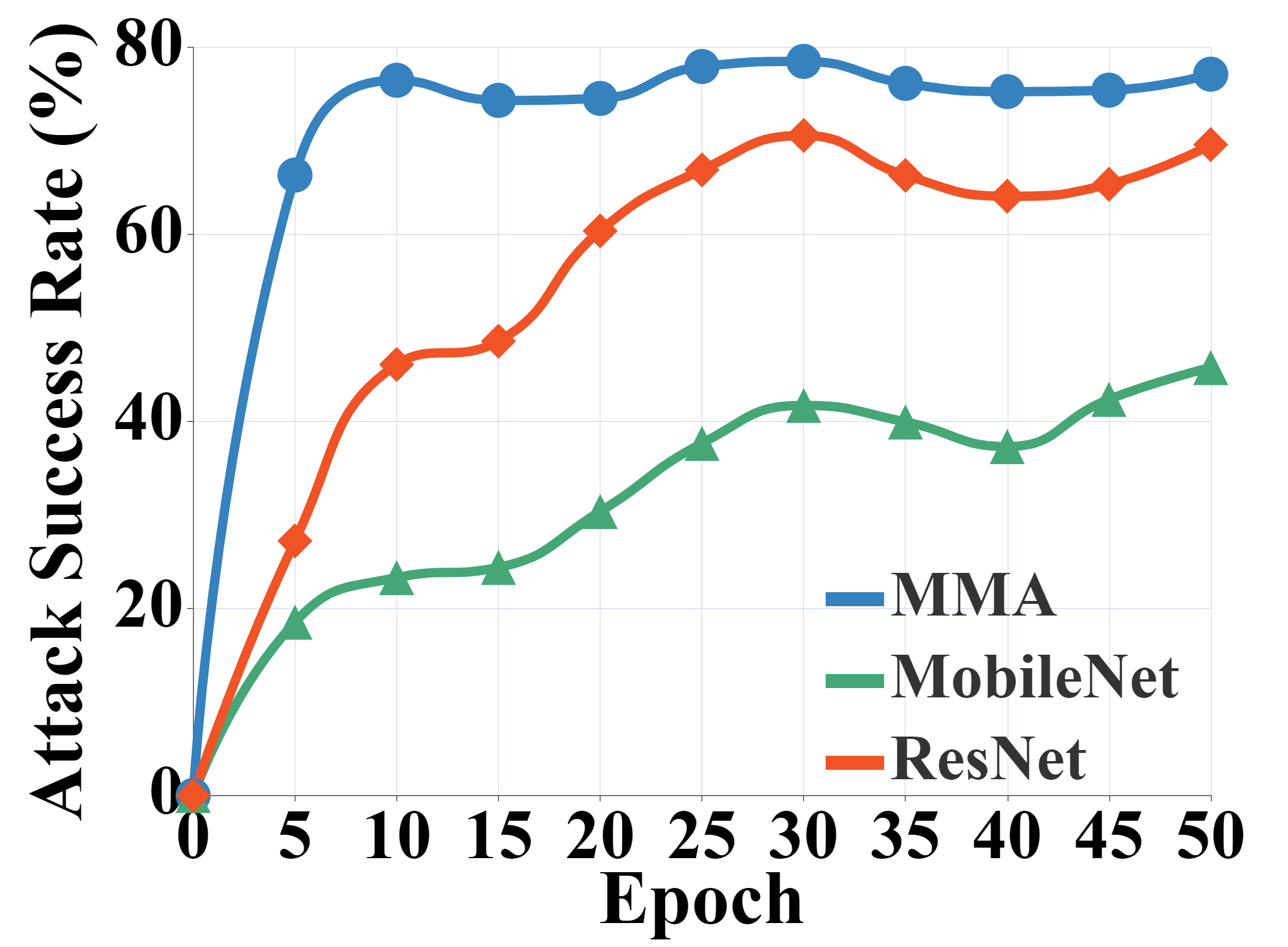}}
    \subfigure[MobileNet]{\includegraphics[width=0.32\linewidth]{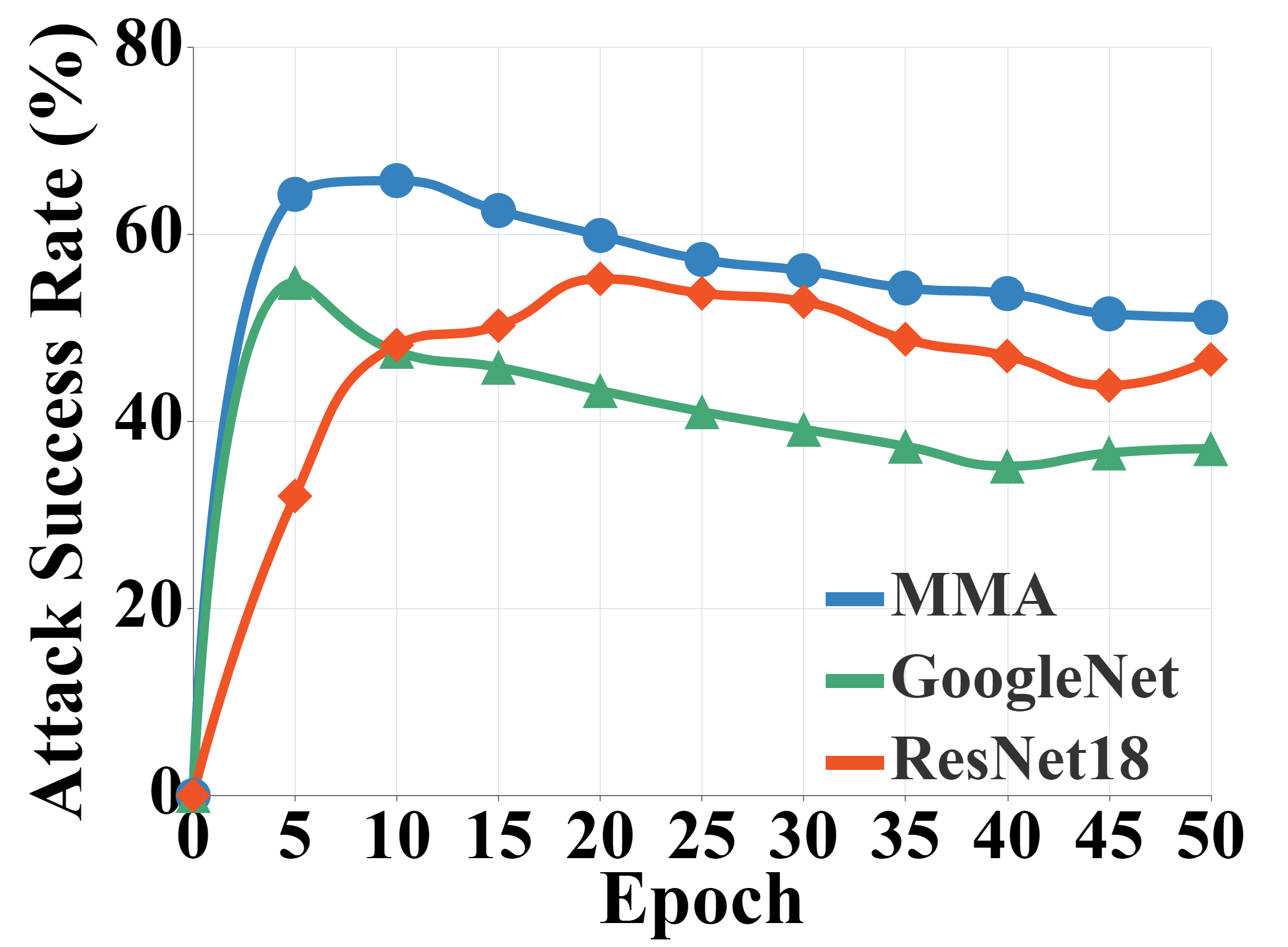}}
    \subfigure[ResNet]{\includegraphics[width=0.32\linewidth]{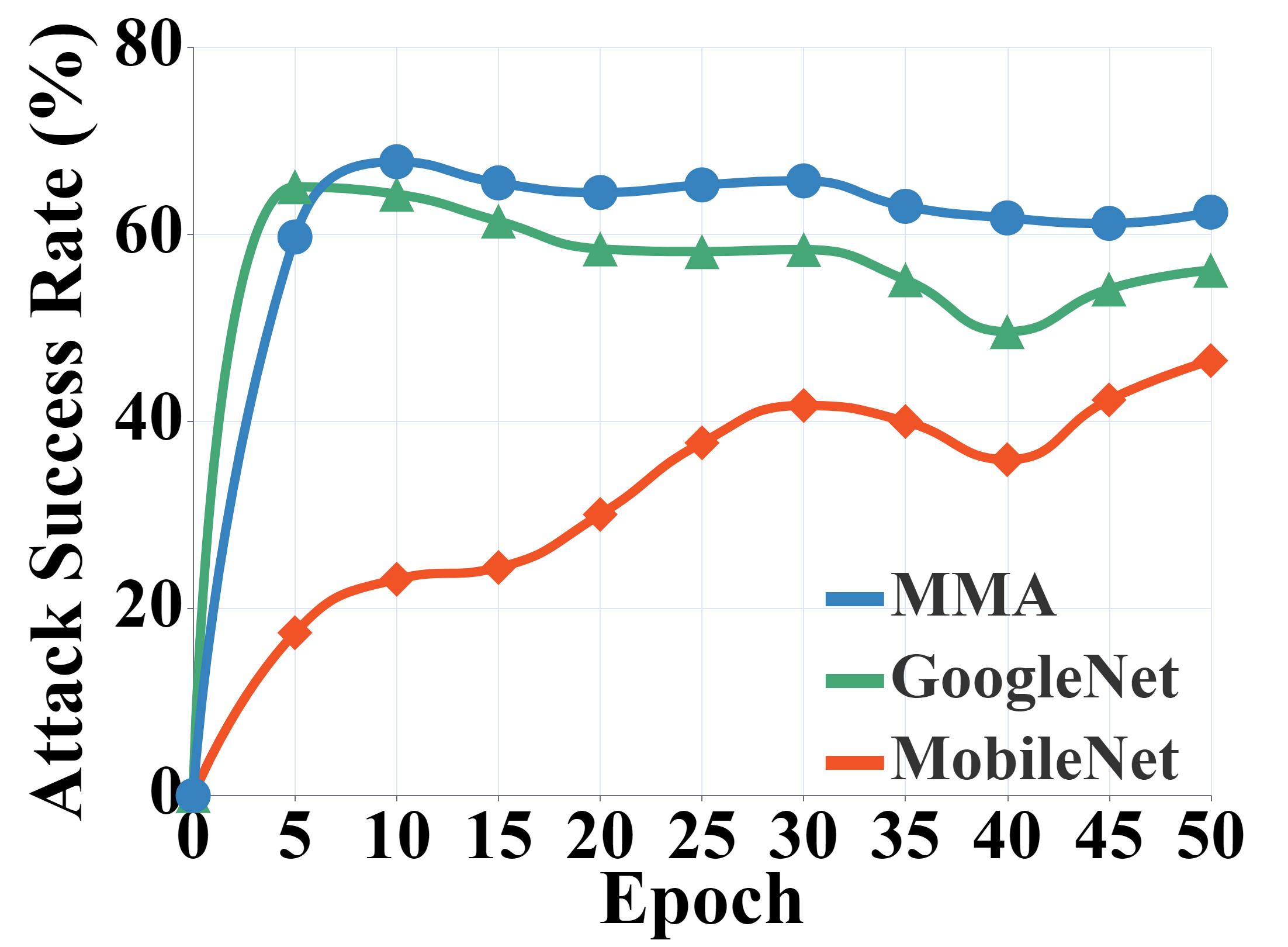}}
    \caption{Multi-step attack performances over training epochs. The size of MAA is fixed to $3 \times 4$.}
    \label{epoch_for_multi}
\end{figure}

\textbf{One-step and multi-step transferability comparison.}
To validate the effectiveness of MMA, Table \ref{singlestep_comp} and Table \ref{multistep_comp} reported the performance of different proxy model architectures against three black-box models using FGSM and BIM, respectively.
For fixed hyperparameters, the epoch and the size\footnote{The $x \times y$ model size denotes $x$ rows and $y$ columns model architecture where each element is a block. In this paper, the standard residual block of ResNet18 is adopted for fair comparisons.} of MMA are fixed into 30 and $3 \times 4$ while the tuned results denotes the best performance of MMA over $\{2,3,4,5\}\times\{2,3,4,5\}$ sizes and $\{5,10,\cdots,50\}$ epochs (the best performance of other proxy models over $\{5,10,\cdots,50\}$ epochs).
Overall, the MMA model architecture consistently yields more transferable AEs.
Specifically, for the single-step attack, MMA can notably improve the transferability of AEs by 2.32\% to 17.96\%.
Furthermore, such improvement is more striking in multi-step attack settings, where AEs based on MAA can achieve a maximum of 79.79\% ASR (from at least 5\% improvement to 40\% improvement).
\par
\begin{figure}[t]
    \centering
    \includegraphics[width=0.6\linewidth]{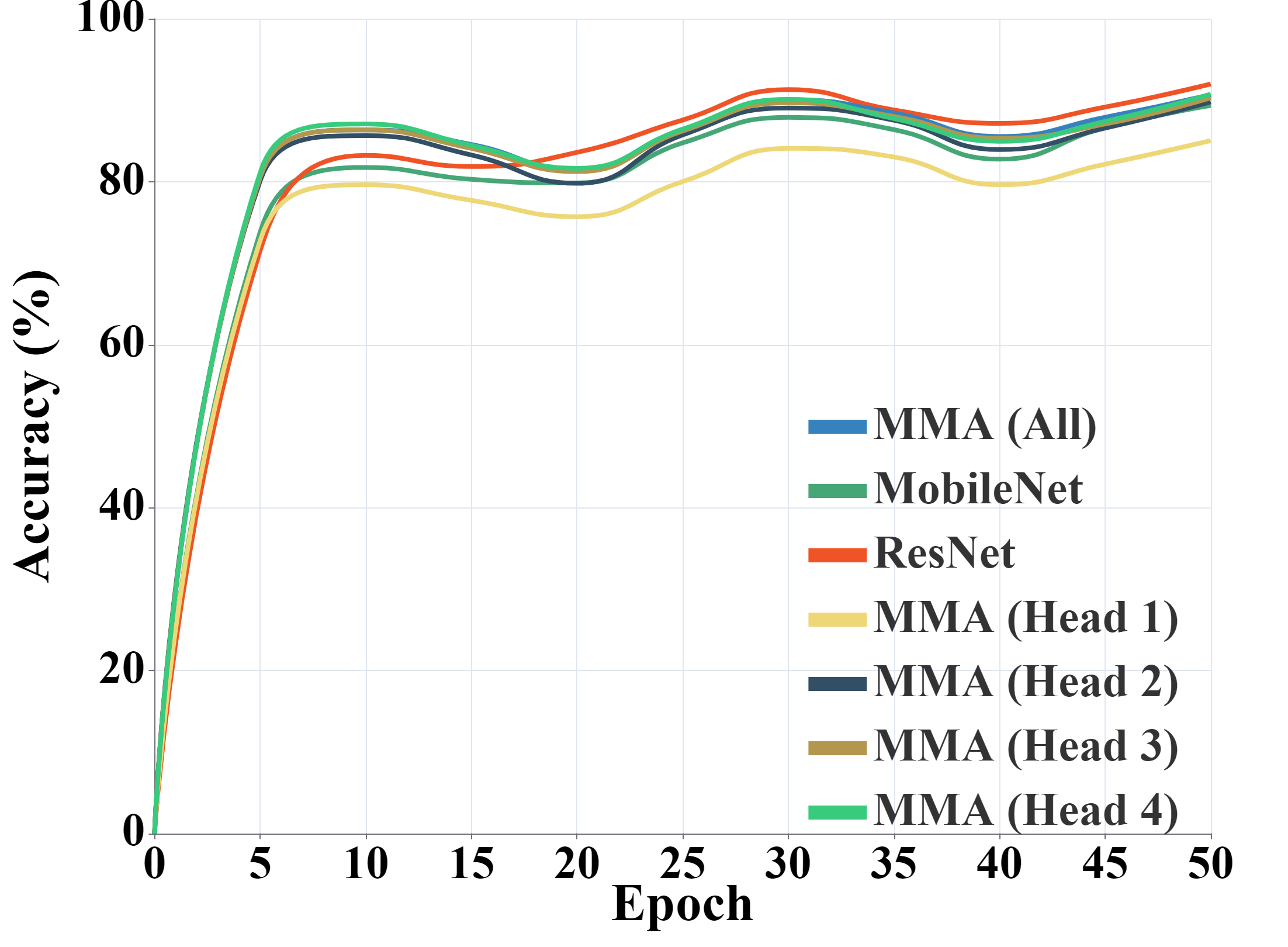}
    \caption{Accuracy of proxy models with different epochs. The size of MAA is fixed to $3 \times 4$.}
    \label{acc}
\end{figure}

\textbf{Epoch impact.}
We are also interested in impact of training epoch to transferability of AEs and Figure \ref{epoch_for_single},\ref{epoch_for_multi},\ref{acc} demonstrated the related results.
Firstly, it is observed that the ASR steadily rises in earlier epochs until its peak of just around the 10-th epoch, and then the ASR starts to fluctuate considerably.
Meanwhile, the varying trend of accuracy in Figure \ref{acc} also is fairly similar to the varying trend of ASR.
In fact, the booming accuracy indicates that the learned low-level feature extractors of the proxy model increasingly align with black-box models, which benefits the transferability of crafted AEs.
Moreover, the weak fluctuation of accuracy in late epochs ($\geq 10$) suggests that the model starts to learn exclusive high-level features to enhance the performance of the model, which weakens the transferability of AEs.

\par
\begin{figure}
    \centering
    \subfigure[Single-Step]{\includegraphics[width=0.45\linewidth]{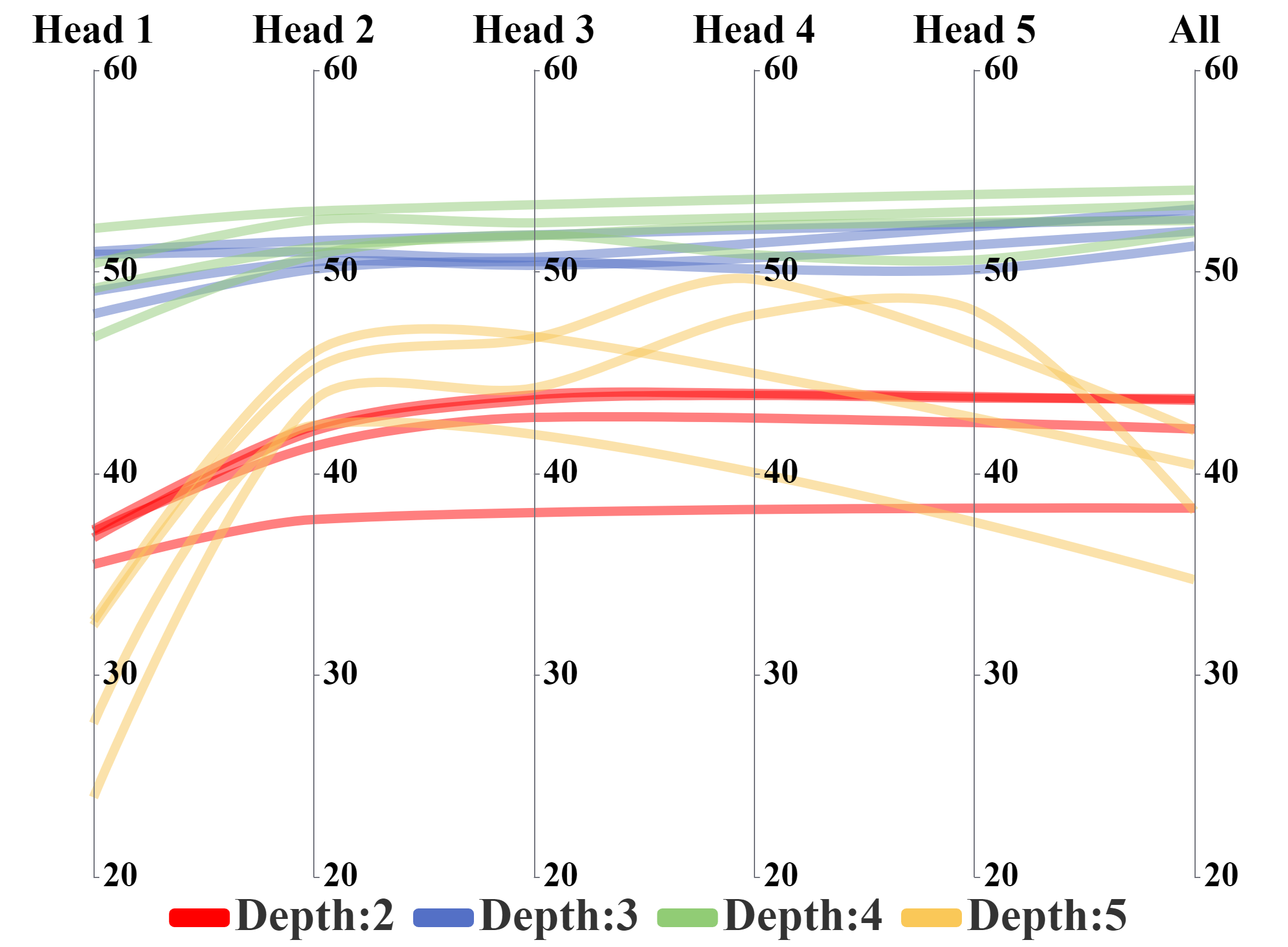}}
    \subfigure[Multi-Step]{\includegraphics[width=0.45\linewidth]{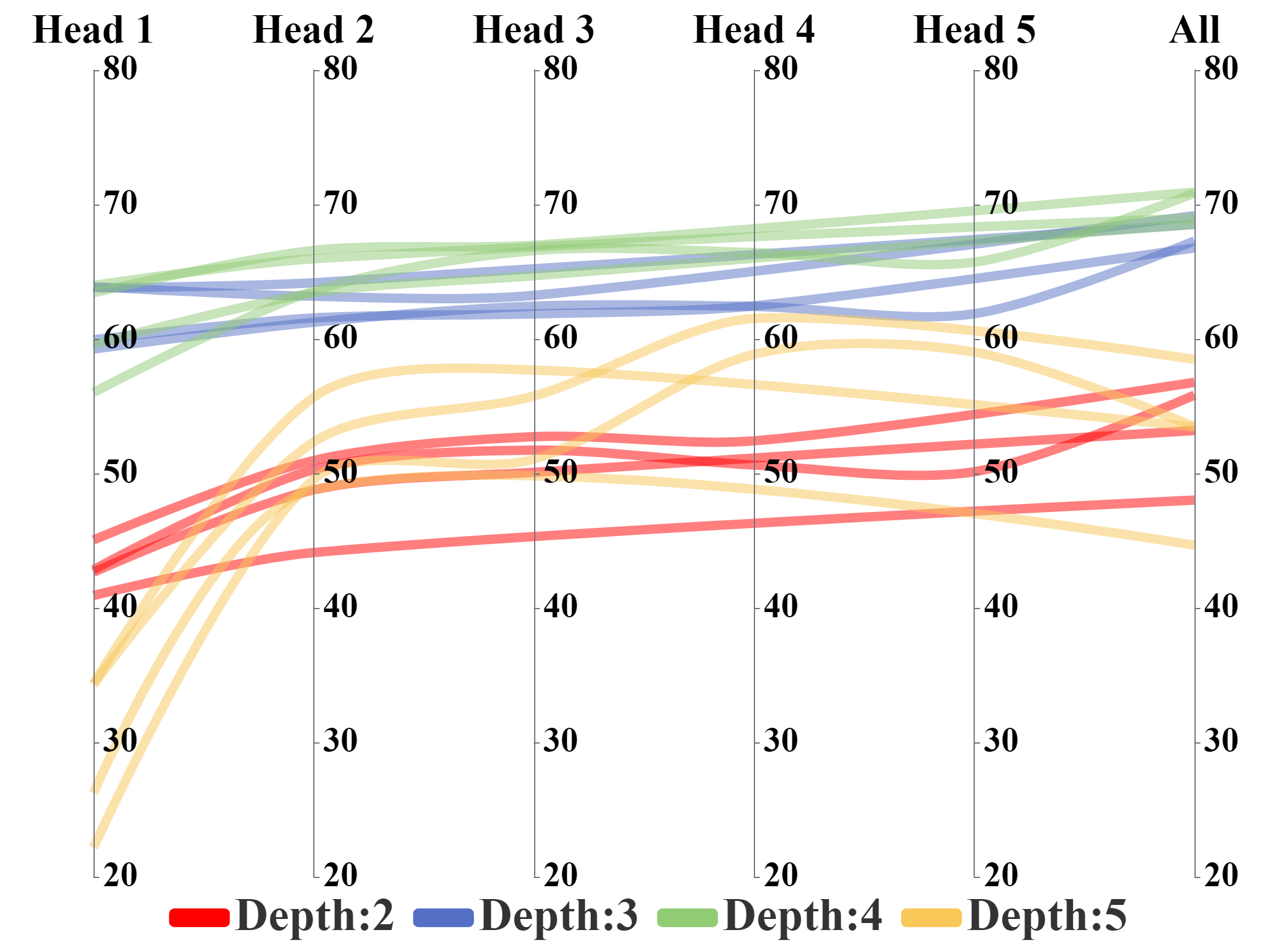}}
    \caption{Attack performance of MMA over different depths and heads. The training epoch is fixed to 30.}
    \label{height_impact}
\end{figure}

\textbf{Model size impact.}
As shown in the earlier section, the size of MAA is a critical factor to the transferability of generated AEs, and Figure \ref{height_impact} and Table \ref{width_impact} illustrate the attack performance over varying MAA depths and widths, respectively.
The transferability of resultant AEs, as demonstrated in Figure \ref{height_impact}, is reached the crest by increasing the depth to 3, and subsequently, an increase of depth will damage the transferability.
The phenomenon is consistent with the earlier deduction that increasing the depth only within a proper range can enhance transferability.
In other words, the depth of 3 is the best trade-off point where the model can greatly learn the low-level model-cross shared features while making disturbance of high-level exclusive features minimal, in our cases.
Likewise, Table \ref{width_impact} also suggests the same conclusion.

\begin{table}[]
\centering
\caption{Attack performance of MMA over different widths and heads. The training epoch is fixed to 30.}
\resizebox{0.40\textwidth}{!}{%
\begin{tabular}{@{}c|c|cccccc@{}}
\toprule
Attack Method & Width & Head 1 & Head 2 & Head 3 & Head 4 & Head 5 & All \\ \midrule
\multirow{4}{*}{Single-Step} & 2 & 41.56 & 46.17 & - & - & - & 44.92 \\
 & 3 & 42.78 & 47.72 & 48.18 & - & - & 47.26 \\
 & 4 & 41.97 & 47.30 & 48.18 & 49.15 & - & 47.58 \\
 & 5 & 38.86 & 46.67 & 47.56 & 48.18 & 48.13 & 46.27 \\ \midrule
\multirow{4}{*}{Multi-Step} & 2 & 48.74 & 55.77 & - & - & - & 57.55 \\
 & 3 & 51.09 & 58.56 & 59.50 & - & - & 61.69 \\
 & 4 & 49.73 & 57.08 & 58.77 & 60.61 & - & 62.64 \\
 & 5 & 45.08 & 56.17 & 57.96 & 59.59 & 59.19 & 61.87 \\ \bottomrule
\end{tabular}}
\label{width_impact}
\end{table}

\begin{figure}
    \centering
    \includegraphics[width=0.38\textwidth]{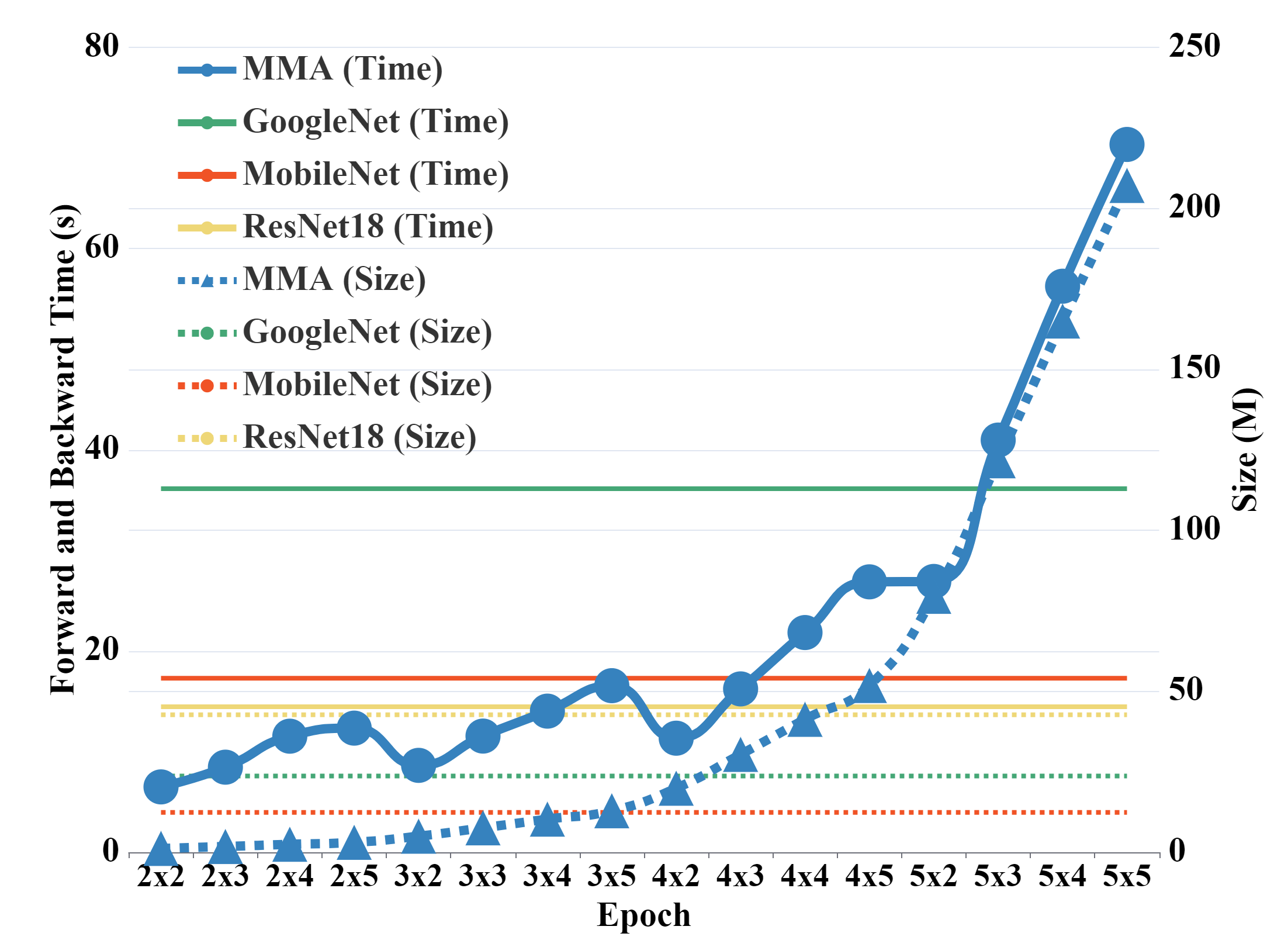}
    \caption{Comparison of costs for launching black-box attacks over different proxy models.}
    \label{overhead}
\end{figure}

\par
\textbf{Attack cost comparsion.}
For practical attacks, the overhead of launching attacks also is crucial, and Figure \ref{overhead} demonstrates the cost to different proxy model architecture.
Noting that the sum time of forward and backward propagation not only can measure the training overhead but also can measure the overhead of crafting AEs.
In Figure \ref{overhead}, the $3\times 4$ is a critical point that denotes whether the overhead of MMA surpasses other baseline models.
We note that in this case, the performance of MMA still significantly surpasses the other baseline models (shown in Table \ref{singlestep_comp} and Table \ref{multistep_comp}).
Therefore, the MMA is comprehensively better than other model architectures, i.e., higher attack performance with fewer or the same costs.


%% file: tex/conclusion.tex
In this work, we proposed enhancing the transferability of AEs from a model architecture perspective and designed multi-track model architecture (MMA) based on deep analysis.
The MMA can adaptively adjust the model size to learn model-cross shared feature extractors under best work with least introduced model-specified feature extractors.
Consequently, the synthesized AEs were maximumly guided towards the vulnerable directions shared by various models, and enjoyed the significant transferability. 
Moreover, we highlighted that the overhead of MMA is comparable to state-of-the-art model architectures and that MMA is a generic technique that can be effortlessly combined with other advanced technologies.
Finally, we conducted extensive experiments to show the effectiveness of MMA.